\documentclass[twoside,leqno,twocolumn]{article}

\usepackage[letterpaper]{geometry}

\usepackage{ltexpprt}
\usepackage{hyperref}
\usepackage{cite}
\usepackage{amsmath,amssymb,amsfonts}
\usepackage{algorithmic}
\usepackage{graphicx}
\usepackage{textcomp}
\usepackage{xcolor}
\usepackage{booktabs}
\usepackage{adjustbox}
\usepackage{subcaption}
\usepackage[switch]{lineno}
\usepackage{newfloat}
\usepackage{listings}
\usepackage{pifont}
\usepackage[switch]{lineno}
\usepackage{soul}

\usepackage{algorithm}

\begin{document}

\newcommand\relatedversion{}
\renewcommand\relatedversion{\thanks{The full version of the paper can be accessed at \protect\url{https://arxiv.org/abs/1902.09310}}} 

\title{\Large VSFormer: Value and Shape-Aware Transformer with Prior-Enhanced Self-Attention for Multivariate Time Series Classification}
\author{
Wenjie Xi\footnotemark[1] \and Rundong Zuo\footnotemark[2] \and Alejandro Alvarez\footnotemark[1] \and Jie Zhang\footnotemark[1] \and Byron Choi\footnotemark[2] \and Jessica Lin\footnotemark[1]
}

\date{}

\maketitle
\renewcommand{\thefootnote}{\fnsymbol{footnote}}
\footnotetext[1]{Department of Computer Science, George Mason University, United States. \{wxi, aalvar10, jzhang7, jessica\}@gmu.edu.}
\footnotetext[2]{Department of Computer Science, Hong Kong Baptist University, Hong Kong. \{csrdzuo, bchoi\}@comp.hkbu.edu.hk.}






\begin{abstract} \small\baselineskip=9pt Multivariate time series classification is a crucial task in data mining, attracting growing research interest due to its broad applications. While many existing methods focus on discovering discriminative patterns in time series, real-world data does not always present such patterns, and sometimes raw numerical values can also serve as discriminative features. Additionally, the recent success of Transformer models has inspired many studies. However, when applying to time series classification, the self-attention mechanisms in Transformer models could introduce classification-irrelevant features, thereby compromising accuracy. To address these challenges, we propose a novel method, VSFormer, that incorporates both discriminative patterns (\textbf{shape}) and numerical information (\textbf{value}). In addition, we extract class-specific prior information derived from supervised information to enrich the positional encoding and provide classification-oriented self-attention learning, thereby enhancing its effectiveness. Extensive experiments on all 30 UEA archived datasets demonstrate the superior performance of our method compared to SOTA models. Through ablation studies, we demonstrate the effectiveness of the improved encoding layer and the proposed self-attention mechanism. Finally, We provide a case study on a real-world time series dataset without discriminative patterns to interpret our model. \end{abstract}

\section{Introduction}\label{intro}

A multivariate time series (MTS) consists of multiple time-ordered measurements or observations across different variables. Multivariate Time Series Classification (MTSC) is one of the key tasks for MTS, and it is widely applied in various fields such as motion recognition \cite{rakthanmanon2013fast}, solar flare prediction \cite{SF2}, and human activity recognition \cite{wang2024hardenvr}. As a result, it has garnered increasing research interest recently \cite{ruiz2021great}. 

\begin{figure}[t]
    \centering\includegraphics[width=0.48\textwidth]{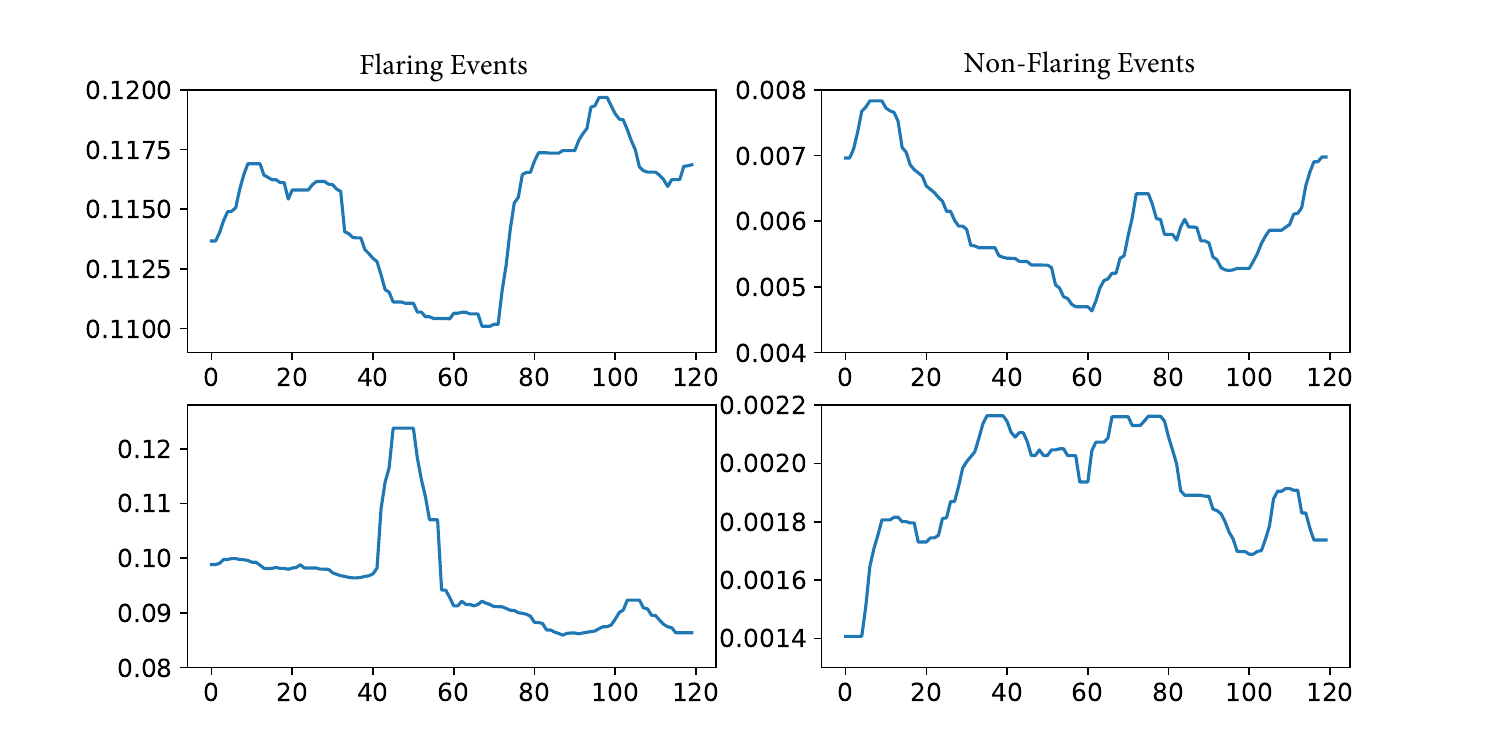}
    \caption{Two time series samples of flaring events (left) and two samples of non-flaring events (right).}
    \label{fig:SF_ex}
\end{figure}

While numerous methodologies have emerged showcasing promising results \cite{ruiz2021great}, one popular kind of approach revolves around identifying and exploiting repeated patterns (subsequences) in time series, treating them as discriminatory features for classification. Such subsequences typically undergo normalization to underscore their shape characteristics. In \cite{shapelets}, the authors introduced the notion of Shapelets---the subsequences that can best represent a class---and use them with traditional classifiers such as decision trees to classify time series. Building upon this idea, recent deep learning methods \cite{shapenet,rlpam,SVPT} have targeted the search for such patterns through the use of neural networks resulting in improved performances.

However, a crucial limitation of these methods is their foundational assumption for the presence of discriminative patterns for MTSC. Many real-world time series data lack such specific patterns that can effectively distinguish one sequence from another, rendering these methods sub-optimal. 
A related example as presented in Figure \ref{fig:SF_ex} is found in solar flare prediction. The two time series on the left correspond to conditions leading to flaring events and those on the right to non-flaring events. While it may be challenging to identify discriminative patterns (also known as \textbf{shape}), the raw numerical information (the \textbf{value}), differing by orders of magnitude, provide a significant distinction between the two classes (The values from the first class (0.08-0.12) are at least one order of magnitude larger than those from the second class (0.0014-0.008)). Shape-based methods require the normalization of subsequences for meaningful comparison of shapes; however, such a step also strips away valuable information presented in the raw values of time series.


Second, the recent success of Transformers \cite{transformer,ViT} has motivated researchers to explore their potential in time series classification \cite{TSTCC,tst,TARNet,SVPT}. However, the self-attention mechanisms of existing Transformer models for MTSC are not classification-oriented. Given they directly make use of time points or subsequences as input, this can result in the introduction of a substantial number of classification irrelevant features~\cite{cheng2023formertime,SVPT} affecting the training of the Transformer due to noise and thus compromising their efficacy for classification.

As a way to address these challenges, we propose the Value and Shape-aware Transformer (VSFormer) with Prior-Enhanced Self-Attention. VSFormer incorporates both discriminative patterns (termed as $\textbf{shape}$) and numerical information (termed as $\textbf{value}$), which enhances the performance in cases where discriminative patterns are lacking, or where both shape and value information are important. Specifically, our model has two branches. The first focuses on learning from the $\textbf{shapes}$ in the time series, while the second extracts insights from the raw $\textbf{values}$. The shape branch locates shape tokens from each time series via motif discovery to find repeated patterns by class, allowing for more targeted and accurate extraction of significant shapes. The value branch then partitions the time series into segments across different granularity levels and calculates interval-based statistics to form value tokens. Moreover, class-specific prior information is extracted for each token and used to enrich the encoding process. Such prior information is also used to provide classification-oriented self-attention learning, enhancing classification-relevant features and thus attenuating irrelevant noise. Finally, a decision layer is designed to fuse shape and value representations for the final classification.

To summarize, our work has the following main contributions: \\
1). We propose VSFormer, a novel approach incorporating both discriminative shapes and numerical information for time series classification.\\
2). We improve the existing positional encoding and introduce class-specific prior information derived from the training data for both shape and value to enrich it. \\
3). We propose Prior-Enhanced Self-Attention which is classification-oriented to enhance classification-relevant features and reduce the impact of noise.\\
4). Extensive experiments were conducted on all 30 UEA archive datasets showing that our model outperforms SOTA models. We also experimented with a real-world application showing the superiority of our model in a case where no discriminative shapes are present.
\section{Related Work}

\subsection{Multivariate Time Series Classification}
Existing works for MTSC can be roughly categorized into distance-based, pattern-based, and deep-learning-based methods. Distance-based methods rely on measuring the dissimilarity between time series using distance measures such as Euclidean Distance (ED) and Dynamic Time Warping (DTW) \cite{DTW} and use 1-nearest-neighbor for classification. Pattern-based methods extract bag-of-patterns or discriminative patterns from raw time series. A prominent example of using bag-of-patterns is WEASEL+MUSE \cite{weasel+muse}, which constructs a multivariate feature vector from each variable of the MTS using various sliding windows, extracts discrete features, and undergoes feature selection to remove non-discriminative features. Meanwhile, several approaches have been devoted to using shapelets \cite{shapelets}, such as Generalized Random Shapelet Forests (gRSF) \cite{gRSF} that generate shapelet-based decision trees by randomly choosing a subset of shapelets.

More recently, deep learning models have shown notable success in MTSC\cite{ruiz2021great}. MLSTM-FCNs \cite{MLSTM-FCN} employ a combination of LSTM and stacked CNN layers for feature extraction from time series. TapNet \cite{tapnet} further introduces the attentional prototype learning for fully and semi-supervised MTSC. ROCKET \cite{rocket} and its optimized counterpart, MiniRocket \cite{minirocket}, use random convolutional kernels for time series transformation and subsequently train classifiers on these transformed features. 

Several deep-learning models also focus on finding discriminative patterns. For instance, ShapeNet \cite{shapenet} utilizes embedding learning to map subsequences into a unified space and employs clustering to discern these patterns. RLPAM \cite{rlpam} uses reinforcement learning to detect patterns beneficial for classification tasks. Meanwhile, SVP-T \cite{SVPT} uses k-means to capture vast amounts of time series subsequences and feed them into a Transformer encoder. Despite their remarkable performances, these models have the assumption that some discriminative patterns exist in the time series, and overlook scenarios where such patterns are absent in the datasets.

\subsection{Transformers for Time Series Classification}
The success of the Transformer model has inspired numerous researchers to adapt it for time series classification. Beyond SVP-T, which is designed specifically for MTSC, several representation learning methods have taken classification as their downstream task. For example, TS-TCC \cite{TSTCC} facilitates unsupervised representation learning from time series data by utilizing temporal and contextual contrasting modules. TST \cite{tst} introduces a novel transformer encoder-based framework for representation learning, while TARNet \cite{TARNet} integrates a task-aware reconstruction strategy to bolster downstream task performance. However, these models introduce many classification-irrelevant features into self-attention learning, potentially compromising the efficacy of time series classification.
\section{Problem Formulation}
A multivariate time series, $\mathbf{X} = \{X^1,...,X^V\} \in \mathbb{R}^{V \times T}$, is a collection of several univariate time series, $X^i \in \{x_1,...,x_T\}$. $T\in \mathbb{Z}^+$ is the total number of observations, and $V\in\mathbb{Z}^+$ represents the number of variables.

An MTS dataset consists of pairs of MTS and associated labels, which can be represented as $\{(\mathbf{X}_1, y_1),...,(\mathbf{X}_N, y_N)\}$. Here, $y = \{y_1,...,y_C\}\in\mathbb{R}^C$ denotes the corresponding class, and $N$ is the total count of MTS instances in the dataset.

A multivariate time series classification problem aims to train a classifier that can predict the class label for an unlabeled, previously unseen MTS.

\section{Methodology}

\begin{figure}[t]
    \centering
    \includegraphics[width=0.48\textwidth]{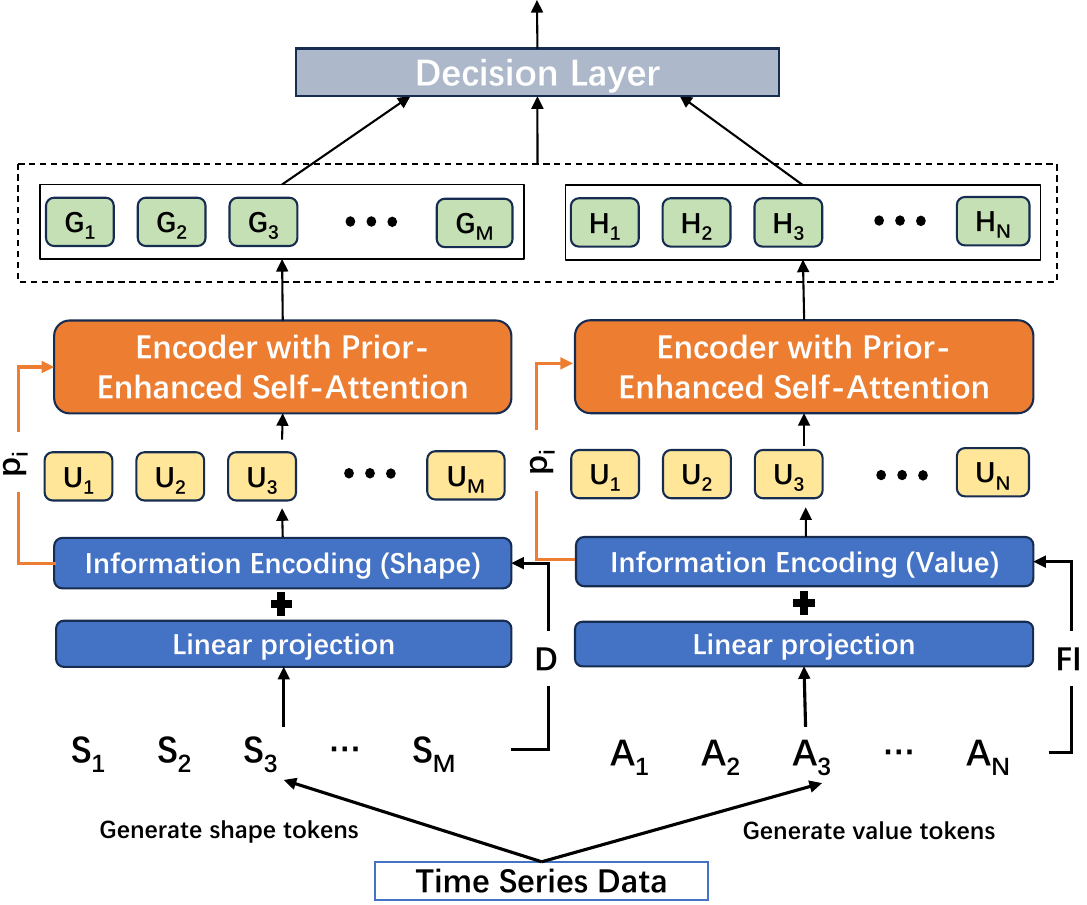}
    \caption{The overall architecture of VSFormer}
    \label{fig:overview}
\vspace{-3mm}
\end{figure}

\subsection{Overall Architecture}
Figure \ref{fig:overview} shows the overall structure of VSFormer. The time series dataset is initially subjected to two distinct preprocessing steps for \textbf{shape} and \textbf{value}, yielding input tokens and class-specific prior information. These tokens are then encoded using our improved Time Series Information (TSI) Encoding and fed into the Transformer encoders with our proposed Prior-Enhanced Self-Attention (PESA). Finally, the representations from the two encoding branches are fused by the decision layer obtaining the final results.

\subsection{Input Token Generation}
In this section, we outline the data preprocessing steps for generating both shape and value tokens.\\

\begin{figure}[t]
    \centering
    \includegraphics[width=0.48\textwidth]{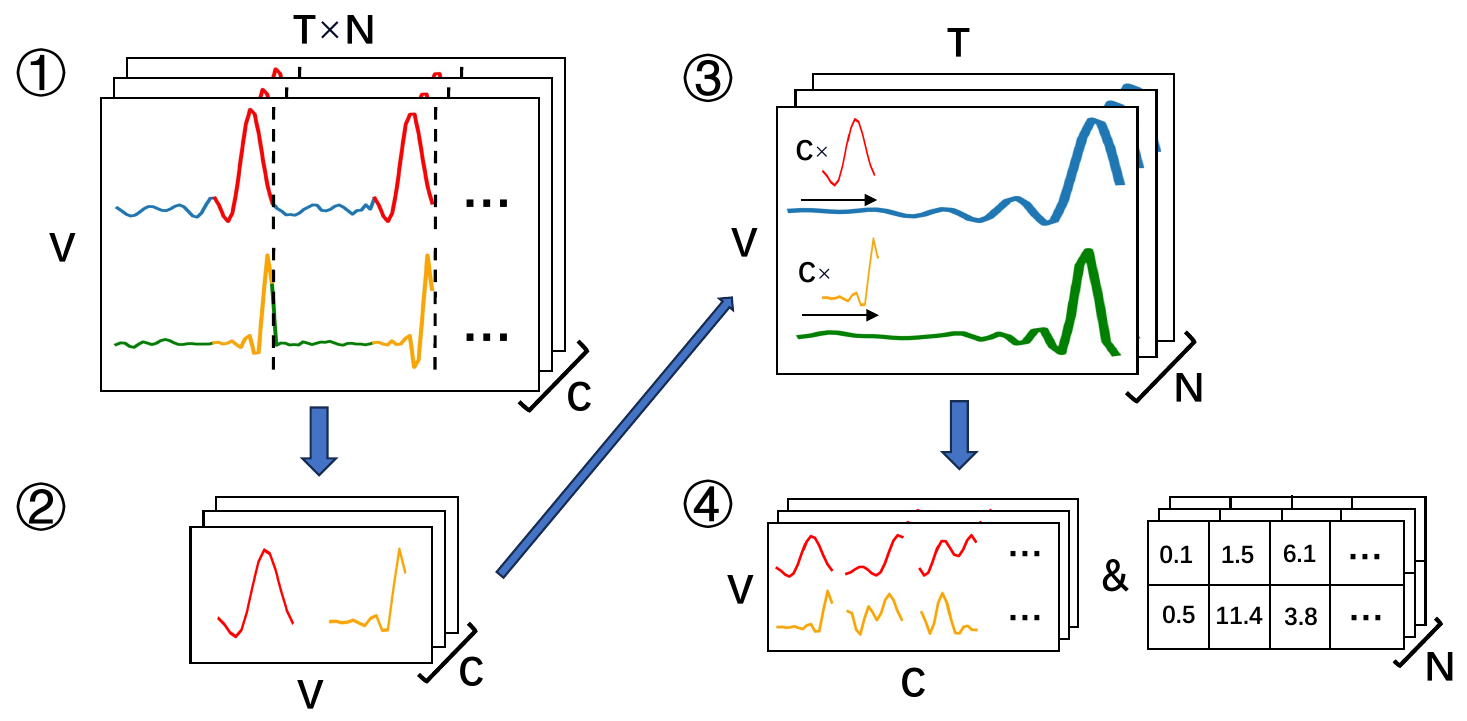}
    \caption{The shape token generation process (with $k=1$ for clarity). Steps: \ding{172} Concatenate sequences per class by variable and identify repeated patterns (highlighted in red and orange). \ding{173} Extract prototype shapes. \ding{174} Conduct a similarity search for each instance. \ding{175} Generate a set of shapes alongside their associated distances.}
    \label{fig:motif}
\vspace{-3mm}
\end{figure}

\noindent\textbf{Shape Tokens.}
Figure \ref{fig:motif} shows the shape token generation process. We employ STOMP \cite{stomp}, a well-known motif (repeated patterns) discovery algorithm to capture significant shapes within the data. All time series from the same class are concatenated by their variable, which results in a lengthy multivariate sequence represented as $\mathcal{X}^{v,c} = \{\mathbf{X}_1^{v,c},..., \mathbf{X}_N^{v,c}\}$, where $v$ denotes the variable, and $c$ denotes the class. Subsequently, we search for the top-$k$ motif pairs within this extended sequence (for illustrative clarity, in Figure \ref{fig:motif}, Step 2, we set $k=1$). This results in a total of $\mathcal{M} = k\times V \times C$ motif pairs. From each motif pair, we choose one motif instance, denoted by $\hat{S}^{v,c}$, as the prototype shape specific to its variable and class. The intuition is to identify a small set of patterns (shapes) representing each variable per class. 
Next, we search within each time series instance in the corresponding variable to locate subsequences similar to these prototype shapes (Figure \ref{fig:motif}, Step 3). Thus, for the $i$-th instance, we acquire shape token set $\mathbf{S}=\{S^{1,1}, S^{1,2},..., S^{V,C}\}$ and their associated Z-normalized Euclidean Distances, represented as $\mathbf{D}=\{D^{1,1}, D^{1,2},..., D^{V,C}\}$ (Figure \ref{fig:motif}, Step 4). To simplify notation, these are rewritten as $\mathbf{S}=\{S_1, S_2,..., S_\mathcal{M}\}$ and $\mathbf{D}=\{D_1, D_2,..., D_\mathcal{M}\}$.\\


\noindent\textbf{Value Tokens.}
Individual time points within a time series are insufficient for revealing meaningful information associated with class labels \cite{cheng2023formertime}. To address this problem, instead of utilizing all time points as inputs, we propose to derive value tokens by calculating statistical features from the various time series intervals. The simplest approach is to divide the time series into $w$ equal-length intervals. However, since the optimal choice of $w$ (or equivalently, the length of the interval) is unknown, to ensure capturing a sufficient amount of information from the time series, we propose a multi-granularity approach, by iterating through the different values of $w$ ranging from 1 to $M\in \mathbb{Z}^+$, and segmenting each time series variable, $X^v$, into $M$ sets of equal-length intervals. Each set of intervals represents a different granularity level $\mathcal{G}^w$. Note that when $w=1$, the interval is the whole time series.


For each interval, we calculate three statistical features: mean ($\mu$), standard deviation ($\sigma$), and slope ($\psi$), which is derived from fitting a linear regression to the subsequence. Notably, \cite{tsf,ruiz2021great} have corroborated the efficacy of these features in time series classification. 

Through this method, we generate a set of value tokens for each instance, symbolized as 
$\mathbf{A}=\{A_{1,\mu}^1, A_{1,\sigma}^1, A_{1,\psi}^1,...,$
$ A_{1,\mu}^w, A_{1,\sigma}^w, A_{1,\psi}^w, ..., A_{M,\mu}^M, A_{M,\sigma}^M,A_{M,\psi}^M\}$ that encapsulates crucial statistical information about the different intervals within the time series. For a multivariate time series, we have a total of $\mathcal{N} = V \times 3 \times [(1+M) \times M / 2]$ statistical features. The value token set $\mathbf{A}$ could then be rewritten as $\mathbf{A} = \{A_1, A_2,..., A_\mathcal{N}\}$ for simplicity.

\subsection{Class-Specific Prior Information}
Class-specific prior information corresponds to each input token and can be directly calculated from supervised information. We introduce it to enrich the encoding process and guide self-attention learning to enhance classification-relevant features and reduce noise.\\

\noindent\textbf{Prior information for shape.}
For shape tokens, we use the corresponding distances of the shapes to calculate the weight as their prior information. 

First, we compute the weights for prototype shapes. For each prototype shape $\hat{S}^{v,c}$, we compute two distinct types of distances, $D_1$ and $D_2$. The former, $D_1$, represents the mean intra-class distance for each prototype shape and is computed by averaging all distances within the same class $c$. 
In contrast, $D_2$ describes the mean inter-class distance and is determined by averaging all distances corresponding to different classes. Using both, we calculate a ratio $\hat{D} = \frac{D_2}{D_1 + D_2}$ that quantifies the distinctiveness of the prototype shapes across classes. A value of $\hat{D} > 0.5$ indicates that the shape is discriminative and significantly contributes to classification. Conversely, $\hat{D} = 0.5$ implies that the shape is common across all classes and does not contribute to discrimination. Note that $\hat{D} < 0.5$ indicates that the prototype shape is a repeated pattern that does not belong to its class, which is obviously contrary to our method and is therefore unlikely to happen.

We then assign a weight $w_{\hat{S}}$ to each prototype shape $\hat{S}^{v,c}$ through the following computations: 
\begin{equation}
    x = \max(\hat{D}-0.5,0),
\end{equation}
\begin{equation}
    w_{\hat{S}} = e^{\alpha x}, \ \alpha \geq 0,
\label{eq:alpha}
\end{equation}
which ensures that prototype shapes with higher discriminative power (larger $\hat{D}$) receive greater weights.

Subsequently, for each shape token $S_i^{v,c}$, we determine its distance $d$ to the prototype shape $\hat{S}^{v,c}$ and assign it a weight $w_{S_i}$ using the formula:
\begin{equation}
    w_{S_i} = \beta e^{-d} + 1, \ \beta \geq 0,
\label{eq:beta}
\end{equation}
which ensures the less similar the shape, the smaller the weight.

Finally, we compute the final weight $w_{\text{final}}^{v,c,i}$, which serves as the prior information for each shape token: 
\begin{equation}
    p_i = w_{\text{final}}^{v,c,i} = w_{\hat{S}^{v,c}} \times w_{S_i^{v,c}}. 
\end{equation}

The procedure ensures that a shape token more similar to a discriminative prototype shape receives a higher weight. Note that $\alpha$ and $\beta$ are weighting hyperparameters. Setting either one to zero will remove the influence of its corresponding weight on the final weights. \\

\noindent\textbf{Prior information for value.}
Generating prior information for value tokens involves an entropy-based feature importance calculation. Given the set of value tokens $\mathbf{A}$ and corresponding class labels from the training set, we first compute the entropy $H(A)$, quantifying the uncertainty associated with these tokens. The entropy is defined as:
\begin{equation}
H(A) = -\sum_{i=1}^{C} Pr(i|A) \log_2 Pr(i|A),
\end{equation}
where $Pr(i|A)$ represents the probability of class $i$ given a value token $A$, and $C$ denotes the total number of classes.

Next, the conditional entropy $H(A|Y)$ measures the average entropy of $A$ given the class label $Y$:
\begin{align}
H(A|Y) &= -\sum_{j=1}^{C} Pr(Y=j) \sum_{i=1}^{C} Pr(i|A, Y=j) \nonumber \\
&\quad \times \log_2 Pr(i|A, Y=j),
\end{align}
where $Pr(Y=j)$ is the probability of class $j$, and $Pr(i|A, Y=j)$ is the probability of class $i$ given a value token $A$ and class label $Y=j$.

The feature importance, denoted by $FI(A)$, is computed using the gain in entropy, which indicates the reduction in uncertainty about class labels provided by the value token $A$:
\begin{equation}
FI(A) = H(A) - H(A|Y).
\end{equation}

The feature importance $FI(A)$ is then used as the prior information for the value token $A$, where higher feature importance indicates a greater relevance of the value token for the classification task.

\subsection{Time Series Information Encoding} \label{TSI_encoding}
While the relative position of shapes within a time series instance is often more crucial than sequence order, the authors of SVP-T\cite{SVPT} proposed a positional encoding scheme to encode the variable ID, start timestamp, and end timestamp of a shape, with all information normalized to the range of [0, 1] (by dividing them with the number of variables or time series length). This approach, however, exhibits a limitation: While variable information is inherently discrete, normalization transforms it into a continuous value, leading to artificial ordering. For instance, with three variables, the positional encoding scheme normalizes them to 1/3, 2/3, and 1, respectively. As a result, the third variable appears numerically three times greater than the first, which is a misleading representation. To preserve the discreteness of variable information, we implement binary encoding to transform the variables into binary digits $B_i$. We further extend this approach by incorporating class-specific prior information into the encoding process. As a result, we introduce Time Series Information (TSI) Encoding. The structure of TSI Encoding for the $i$-th token is outlined in Table \ref{encoding}.

\begin{table}[h]
\centering

\caption{Time Series Information Encoding}
\begin{adjustbox}{width=0.46\textwidth,center}
\begin{tabular}{|c|c|c|c|}
\hline
Variable & Start timestamp & End timestamp & Prior \\ \hline
$B_i$  &  $t_{i,start}/T$ &  $t_{i,end}/T$   &  $p_i$  \\ \hline
\end{tabular}
\end{adjustbox}
\label{encoding}
\vspace{-4mm}
\end{table}

\subsection{Encoding Layer}
In the encoding layer, the output from TSI Encoding, $I_i$, is passed through a linear transformation. Simultaneously, the input termed $Token$ is processed by a distinct linear projection layer. The outcomes from both processes are then added to form the input $U_i$ for the Transformer Encoder. The process can be represented as:
\begin{equation}
    U_i = I_i W_I + W_S Token,
\end{equation}
where $W_I \in \mathbb{R}^{d_I \times d_{model}}$ and $W_S \in \mathbb{R}^{d_{model} \times d_{Token}}$ are trainable weights, $d_{model}$ denotes the input dimension of the Transformer Encoder, while $d_I$ and $d_{Token}$ represent the dimensions of $I_i$ and $Token$, respectively.

\subsection{Prior-Enhanced Self-Attention} \label{PESA}

Here, we introduce a time series classification-oriented self-attention, Prior-Enhanced Self-Attention (PESA) mechanism, which incorporates class-specific prior information into the attention computation to enhance classification-relevant features and attenuate noise. 

As presented by \cite{transformer}, the conventional self-attention mechanism employs the query matrix $Q$, key matrix $K$, and value matrix $V$. The attention score matrix $A$ is given by:
\begin{equation}
A = \text{softmax}\left(\frac{QK^T}{\sqrt{d}}\right),
\end{equation}
where $d$ is the dimensionality of the queries and keys.

We bring in the prior score matrix $P$, constructed using class-specific prior information $p_i$. Specifically, every element $P_{i,j}$ of matrix $P$ is characterized as:
\begin{equation}
P_{i,j} =
\begin{cases}
p_i \cdot p_j & \text{if } i \neq j \\
1 & \text{if } i = j.
\end{cases}
\end{equation}

This matrix $P$ effectively captures the interactions among different input tokens, emphasizing those that are highly informative for classification within the attention mechanism. 

The final PESA is deduced by taking an element-wise product (represented by $\otimes$) of the attention score matrix $A$ and the prior score matrix $P$, followed by a softmax operation, and finally multiplying the resulting matrix with the value matrix $V$:
\begin{equation}
\text{PESA}(Q,K,V,P) = \text{softmax}(A \otimes P)V.
\end{equation}

By integrating class-specific prior information, we infuse the supervised knowledge externally, which optimizes the self-attention learning process, enhancing its performance for time series classification.

\subsection{Decision Layer}
The decision layer plays an instrumental role in the fusion of shape and value representations learned from previous network layers. These representations, denoted as $R_{\text{shape}}$ and $R_{\text{value}}$, are transformed into class probability spaces, $G$ and $H$, respectively, through two distinct linear layers:
\begin{equation}
    G = W_{\text{shape}} R_{\text{shape}},
\end{equation}
\vspace{-8mm}
\begin{equation}
    H = W_{\text{value}} R_{\text{value}},
\end{equation}
where $W_{\text{shape}}$ and $W_{\text{value}}$ are the learnable weights for the shape and value representations, correspondingly.

We introduce a balancing factor $\lambda$ to regulate the contribution of shape and value information toward the final decision. This factor is computed by passing the concatenated $R_{\text{shape}}$ and $R_{\text{value}}$ through a linear layer, followed by a sigmoid activation function:
\begin{equation}
\lambda = \sigma(W_{\lambda} [R_{\text{shape}}, R_{\text{value}}]),
\label{eq:lambda_1}
\end{equation}
where $W_{\lambda}$ represents the learnable weight and $\sigma$ is the sigmoid activation function.

The layer then computes the final class probabilities $O$ as a $\lambda$-weighted combination of $G$ and $H$, followed by a softmax operation:
\begin{equation}
O = \text{softmax}(\lambda G + (1-\lambda)H).
\label{eq:lambda_2}
\end{equation}

By learning $\lambda$ from both shape and value representations, the model can adaptively adjust the emphasis on these aspects depending on their relevance to the classification task. That is, if the shape information is crucial, $\lambda$ will be closer to 1, giving higher weight to $G$. Conversely, if the value information is more important, $\lambda$ will approach 0, and the influence of $H$ will be stronger.
\section{Experiments}

\begin{table*}[!ht]
\caption{Results of our method and baseline methods on all 30 UEA archive datasets}
\begin{adjustbox}{width=0.98\textwidth,center}
\begin{tabular}{@{}l|cccccccccccccc@{}}
\toprule
                          & EDI    & DTWI           & DTWD           & \begin{tabular}[c]{@{}c@{}}MLSTM\\ -FCNs\end{tabular} & \begin{tabular}[c]{@{}c@{}}WEASEL\\ +MUSE\end{tabular} & SRL            & TapNet         & ShapeNet       & ROCKET         & MiniRocket     & RLPAM          & TST            & SVP-T           & VSFormer           \\ \midrule
ArticularyWordRecognition & 0.970  & 0.980          & 0.987          & 0.973                                                 & 0.990                                                  & 0.987          & 0.987          & 0.987          & \textbf{0.993} & 0.992          & 0.923          & 0.983          & \textbf{0.993} & \textbf{0.993}\\
AtrialFibrillation        & 0.267  & 0.267          & 0.220          & 0.267                                                 & 0.333                                                  & 0.133          & 0.333          & 0.400          & 0.067          & 0.133          & \textbf{0.733} & 0.200          & 0.400          & 0.467
\\
BasicMotions              & 0.676  & \textbf{1.000} & 0.975          & 0.950                                                 & \textbf{1.000}                                         & \textbf{1.000} & \textbf{1.000} & \textbf{1.000} & \textbf{1.000} & \textbf{1.000} & \textbf{1.000} & 0.975          & \textbf{1.000} & \textbf{1.000}\\
CharacterTrajectories     & 0.964  & 0.969          & 0.989          & 0.985                                                 & 0.990                                                  & 0.994          & \textbf{0.997} & 0.980          & N/A            & 0.993          & 0.978          & N/A            & 0.990          & 0.991
\\
Cricket                   & 0.944  & 0.986          & \textbf{1.000} & 0.917                                                 & \textbf{1.000}                                         & 0.986          & 0.958          & 0.986          & \textbf{1.000} & 0.986          & \textbf{1.000} & 0.958          & \textbf{1.000} & \textbf{1.000}\\
DuckDuckGeese             & 0.275  & 0.550          & 0.600          & 0.675                                                 & 0.575                                                  & 0.675          & 0.575          & \textbf{0.725} & 0.520          & 0.650          & 0.700          & 0.480          & 0.700          & 0.700
\\
EigenWorms                & 0.549  & N/A            & 0.618          & 0.504                                                 & 0.890                                                  & 0.878          & 0.489          & 0.878          & 0.901          & \textbf{0.962} & 0.908          & N/A            & 0.923          & 0.725
\\
Epilepsy                  & 0.666  & 0.978          & 0.964          & 0.761                                                 & \textbf{1.000}                                         & 0.957          & 0.971          & 0.987          & 0.993          & \textbf{1.000} & 0.978          & 0.920          & 0.986          & 0.986
\\
ERing                     & 0.133  & 0.133          & 0.133          & 0.133                                                 & 0.133                                                  & 0.133          & 0.133          & 0.133          & \textbf{0.981} & \textbf{0.981} & 0.819          & 0.933          & 0.937          & 0.970
\\
EthanolConcentration      & 0.293  & 0.304          & 0.323          & 0.373                                                 & 0.430                                                  & 0.236          & 0.323          & 0.312          & 0.380          & 0.468          & 0.369          & 0.337          & 0.331          & \textbf{0.471}\\
FaceDetection             & 0.519  & N/A            & 0.529          & 0.545                                                 & 0.545                                                  & 0.528          & 0.556          & 0.602          & 0.630          & 0.620          & 0.621          & \textbf{0.681} & 0.512          & 0.646
\\
FingerMovements           & 0.550  & 0.520          & 0.530          & 0.580                                                 & 0.490                                                  & 0.540          & 0.530          & 0.580          & 0.530          & 0.550          & 0.640          & \textbf{0.776} & 0.600          & 0.650
\\
HandMovementDirection     & 0.278  & 0.306          & 0.231          & 0.365                                                 & 0.365                                                  & 0.270          & 0.378          & 0.338          & 0.446          & 0.392          & \textbf{0.635} & 0.608          & 0.392          & 0.514
\\
Handwriting               & 0.200  & 0.316          & 0.286          & 0.286                                                 & \textbf{0.605}                                         & 0.533          & 0.357          & 0.452          & 0.585          & 0.507          & 0.522          & 0.305          & 0.433          & 0.421
\\
Heartbeat                 & 0.619  & 0.658          & 0.717          & 0.663                                                 & 0.727                                                  & 0.737          & 0.751          & 0.756          & 0.726          & 0.771          & 0.779          & 0.712          & \textbf{0.790} & 0.766
\\
InsectWingbeat            & 0.128  & N/A            & N/A            & 0.167                                                 & N/A                                                    & 0.160          & 0.208          & 0.250          & N/A            & 0.595          & 0.352          & \textbf{0.684} & 0.184          & 0.200
\\
JapaneseVowels            & 0.924  & 0.959          & 0.949          & 0.976                                                 & 0.973                                                  & 0.989          & 0.965          & 0.984          & 0.965          & 0.989          & 0.935          & \textbf{0.994} & 0.978          & 0.981
\\
Libras                    & 0.833  & 0.894          & 0.870          & 0.856                                                 & 0.878                                                  & 0.867          & 0.850          & 0.856          & 0.906          & \textbf{0.922} & 0.794          & 0.844          & 0.883          & 0.894
\\
LSST                      & 0.456  & 0.575          & 0.551          & 0.373                                                 & 0.590                                                  & 0.558          & 0.568          & 0.590          & 0.639          & 0.643          & 0.643          & 0.381          & \textbf{0.666} & 0.616
\\
MotorImagery              & 0.510  & N/A            & 0.500          & 0.510                                                 & 0.500                                                  & 0.540          & 0.590          & 0.610          & 0.560          & 0.550          & 0.610          & N/A            & \textbf{0.650} & \textbf{0.650}\\
NATOPS                    & 0.850  & 0.850          & 0.883          & 0.889                                                 & 0.870                                                  & 0.944          & 0.939          & 0.883          & 0.894          & 0.928          & \textbf{0.950} & 0.900          & 0.906          & 0.933
\\
PenDigits                 & 0.973  & 0.939          & 0.977          & 0.978                                                 & 0.948                                                  & \textbf{0.983} & 0.980          & 0.977          & 0.982          & N/A            & 0.982          & 0.974          & \textbf{0.983} & \textbf{0.983}\\
PEMS-SF                   & 0.705  & 0.734          & 0.711          & 0.699                                                 & N/A                                                    & 0.688          & 0.751          & 0.751          & 0.832          & 0.522          & 0.632          & \textbf{0.919} & 0.867          & 0.780
\\
Phoneme                   & 0.104  & 0.151          & 0.151          & 0.110                                                 & 0.190                                                  & 0.246          & 0.175          & \textbf{0.298} & 0.280          & 0.292          & 0.175          & 0.088          & 0.176          & 0.198
\\
RacketSports              & 0.868  & 0.842          & 0.803          & 0.803                                                 & \textbf{0.934}                                         & 0.862          & 0.868          & 0.882          & 0.921          & 0.868          & 0.868          & 0.829          & 0.842          & 0.908
\\
SelfRegulationSCP1        & 0.771  & 0.765          & 0.775          & 0.874                                                 & 0.710                                                  & 0.846          & 0.652          & 0.782          & 0.846          & \textbf{0.925} & 0.802          & \textbf{0.925} & 0.884          & \textbf{0.925}\\
SelfRegulationSCP2        & 0.483  & 0.533          & 0.539          & 0.472                                                 & 0.460                                                  & 0.556          & 0.550          & 0.578          & 0.540          & 0.522          & 0.632 & 0.589          & 0.600          & \textbf{0.644}\\
SpokenArabicDigits        & 0.967  & 0.959          & 0.963          & 0.990                                                 & 0.982                                                  & 0.956          & 0.983          & 0.975          & \textbf{0.998} & 0.993          & 0.621          & 0.993          & 0.986          & 0.982
\\
StandWalkJump             & 0.200  & 0.333          & 0.200          & 0.067                                                 & 0.333                                                  & 0.400          & 0.400          & 0.533          & 0.530          & 0.333          & \textbf{0.667} & 0.267          & 0.467          & 0.533
\\
UWaveGestureLibrary       & 0.881  & 0.868          & 0.903          & 0.891                                                 & 0.916                                                  & 0.884          & 0.894          & 0.906          & 0.938          & 0.938          & \textbf{0.944} & 0.903          & 0.941          & 0.909
\\ \midrule
Average rank              & 11.700& 10.650& 10.000& 9.783& 7.750& 7.867& 7.800& 6.300& 6.000& 5.200& 5.600& 8.217& 4.667& \textbf{3.817}\\
Num.Top-1                 & 0
& 1& 1& 0& 5& 2& 2& 3& 5& 6& 7& 6& 7& \textbf{8}\\
Num.Top-3                 & 0
& 2& 1& 0& 6& 6& 3& 6& 13& 15& 15& 8& 13& \textbf{17}\\
Num.Top-5                 & 1
& 2& 1& 5& 10& 8& 8& 14& 18& 20& 19& 10& 21& \textbf{24}\\
Ours 1-to-1 wins/draws    & 30& 30& 30& 29& 25& 24& 26& 23& 19& 17& 20& 23& 23
& -              \\
Wilcoxon Test p-value     & 0.000& 0.000          & 0.000          & 0.000                                                 & 0.001                                                  & 0.001          & 0.000          & 0.006          & 0.056          & 0.175          & 0.286          & 0.005          & 0.046          & -              \\ \bottomrule
\end{tabular}
\end{adjustbox}
\label{results}
\end{table*}

\subsection{Datasets}
We evaluate our method on all 30 datasets from the well-known UEA MTSC archive \cite{UEA}. We also conduct a case study using the extensive \textit{Space Weather Analytics for Solar Flares (SWAN-SF)} dataset \cite{angryk_swan-sf_2020} in Section~\ref{case_study}.


\subsection{Comparison Methods}
We compare our proposed approach with three benchmark methods \cite{UEA} in MTSC: (1) \textbf{EDI}, the 1-nearest neighbor with Euclidean Distance, (2) \textbf{DTWI}, Dimension-Independent Dynamic Time Warping, and (3) \textbf{DTWD}, Dimension-Dependent Dynamic Time Warping. We also compare with ten SOTA methods: (1) \textbf{WEASEL+MUSE} \cite{weasel+muse}, the state-of-the-art bag-of-patterns model which extracts features into word representations, (2) \textbf{SRL} \cite{SRL}, a representation learning approach using negative sampling with an encoder network structure, (3) \textbf{MLSTM-FCNs} \cite{MLSTM-FCN}, a deep learning framework combining an LSTM layer and the FCN layer with Squeeze-and-Excitation mechanism, (4) \textbf{TapNet} \cite{tapnet}, an attentional prototypical network for semi- and fully-supervised learning in MTSC, (5) \textbf{ShapeNet} \cite{shapenet}, a network which projects subsequences into a unified space through embedding learning and uses clustering to find discriminative patterns, (6) \textbf{ROCKET} \cite{rocket}, a method using random convolutional kernels to transform time series and using the transformed features to train classifier, (7) \textbf{MiniRocket} \cite{minirocket}, a streamlined version of ROCKET, offers faster computation by optimizing the feature transformation process without compromising accuracy. (8) \textbf{RLPAM} \cite{rlpam}, the network uses reinforcement learning to find patterns that can provide useful information for classifiers. (9) \textbf{TST} \cite{tst}, a timestamp-level Transformer model which uses per-time-point input mechanism, (10) \textbf{SVP-T} \cite{SVPT}, the SOTA Transformer-based model in MTSC which uses shapes as the input.


\subsection{Experiment Settings}
We conduct our experiments on a machine equipped with an AMD EPYC Rome 7742 CPU @ 2.60 GHz and an NVIDIA Tesla A100 GPU. Our implementation utilizes Python 3.8.6 and Pytorch 2.0.0. For the UEA datasets, we adhere to the original splits for the training set and test set. In the case of the solar flare dataset, we allocate 80\% for training and the remaining 20\% for testing. Across all experiments, we reserve 20\% of the training set for validation, aiding in hyperparameter tuning. Regarding the parameters $\alpha$ and $\beta$ in Formula \ref{eq:alpha} and \ref{eq:beta}, we set $\alpha = 3$ and $\beta = 4$. This configuration ensures that the prototype shape and the shape token have comparable weights ($max(w_{\hat{S}}) \approx max(w_{S_i})$), and make enough final weight difference between optimal and sub-optimal shapes. The details of hyperparameters are shown in Appendix~\ref{ap:hyper}. 
We use classification accuracy as the evaluation metric and report the average rank, the number of top-1, top-3, and top-5 accuracy for each model. We also show how many wins/draws our model has compared to others to show performance superiority. To underscore the statistical significance, we employ the Wilcoxon signed-rank test. 


\subsection{Experimental Results}

All the results for the baseline methods are sourced from the original papers or the survey paper \cite{ruiz2021great}, except for the results of TST, which are sourced from \cite{SVPT}, and those of both ROCKET and MiniRocket, which are obtained from \cite{rlpam}. All the best are highlighted in bold. Any result denoted by "N/A" means that it is not reported in the original paper or cannot be produced.

From Table \ref{results}, we can observe that our proposed method outperforms all baseline methods in terms of accuracy. Specifically, the average rank of our method is 3.817, which is the best among all methods. Moreover, our model leads in top-1, top-3, and top-5 accuracies for 8, 17, and 24 datasets, respectively. Compared with the closest SOTA model, SVP-T, which only use shape as the input, our model wins/ties 23 out of 30 datasets. Additionally, the Wilcoxon signed-rank test indicates that our model significantly outperforms all three benchmark methods, EDI, DTWI, and DTWD, and seven state-of-the-art methods, MLSTM-FCNs, WEASEL+MUSE, SRL, TapNet, ShapeNet, TST and SVP-T, with a significance factor of $p < 0.05$. It is worth noting that TST and SVP-T are two SOTA Transformers in MTSC.

\begin{table}[t]
\caption{Ablation study of each branch in VSFormer}
\begin{adjustbox}{width=0.42\textwidth,center}
\begin{tabular}{@{}lccc@{}}
\toprule
                 & Value only & Shape only & VSFormer \\ \midrule
Average accuracy & 0.643      & 0.693      & \textbf{0.748}    \\
Average rank     & 2.533      & 2.317      & \textbf{1.150}   \\ \bottomrule
\end{tabular}
\end{adjustbox}
\label{ablation}
\vspace{-2mm}
\end{table}

\subsection{Ablation Study}

We explore the impact of distinct components within our proposed method, concentrating on comparing our TSI encoding with learnable positional encoding, and our PESA against the original self-attention. Additionally, we study the effectiveness of combining value and shape information in our model.

\begin{figure}[t]
    \centering
    \begin{subfigure}[b]{0.425\columnwidth}
        \includegraphics[width=1\linewidth]{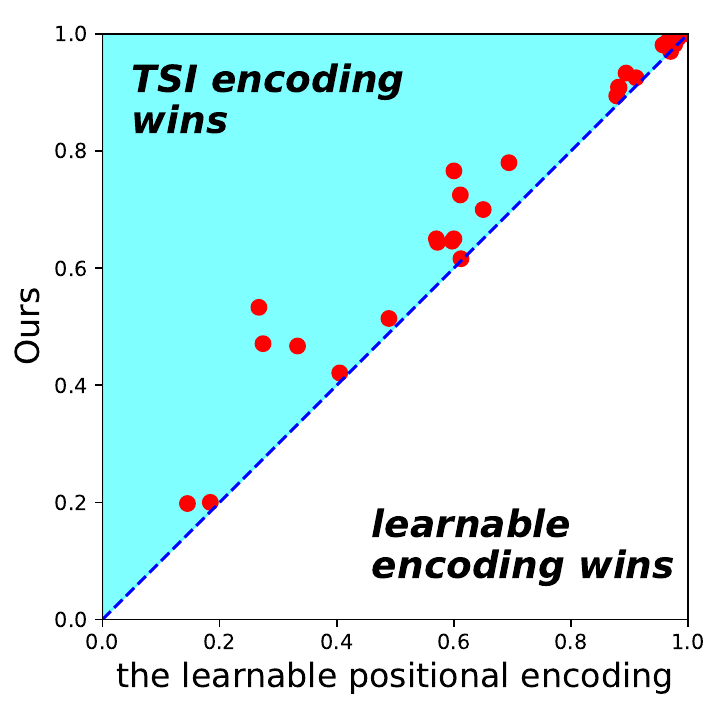}
        \caption{Our method versus the learnable positional encoding.}
        \label{fig:ablation1}
    \end{subfigure}
    \hspace{10pt}
    \begin{subfigure}[b]{0.425\columnwidth}
        \includegraphics[width=1\linewidth]{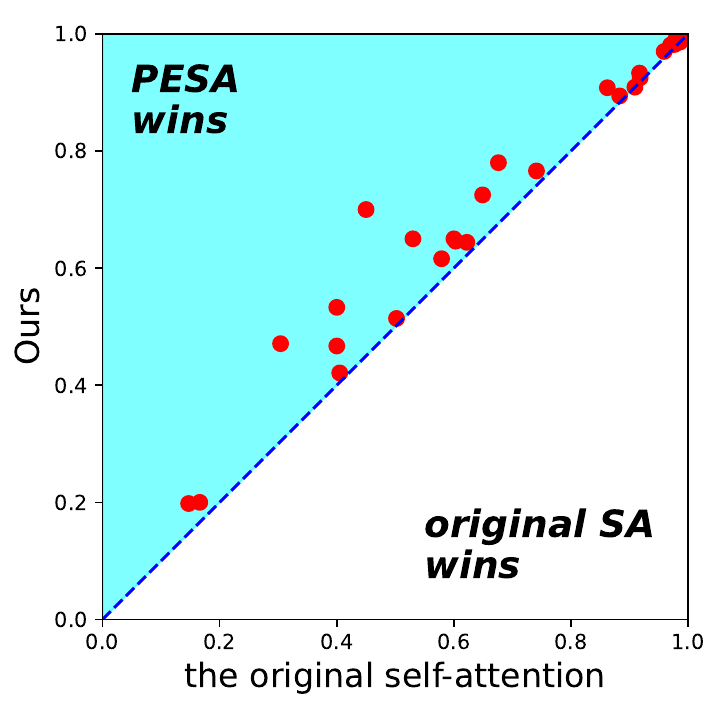}
        \caption{Our method versus the original self-attention.}
        \label{fig:ablation2}
    \end{subfigure}
    \caption{Ablation studies showing the comparative performance of our method with different configurations.}
    \label{fig:ablation}
\vspace{-5mm}
\end{figure}

Initially, we experiment with the learnable positional encoding, in place of our TSI encoding, across all 30 UEA archive datasets. As visualized in Figure \ref{fig:ablation1}, our method wins/draws (as indicated by red dots in the cyan region) in terms of accuracy on all datasets, showing the superiority of our proposed TSI encoding in Section~\ref{TSI_encoding}. Subsequently, we replace our PESA with the vanilla self-attention mechanism \cite{transformer} and conduct the experiments. As depicted in Figure \ref{fig:ablation2}, all red dots reside in the cyan region, further illustrating the effectiveness of our novel self-attention in Section~\ref{PESA} over the original one.

\begin{figure}[t]
    \centering
    \includegraphics[width=0.40\textwidth]{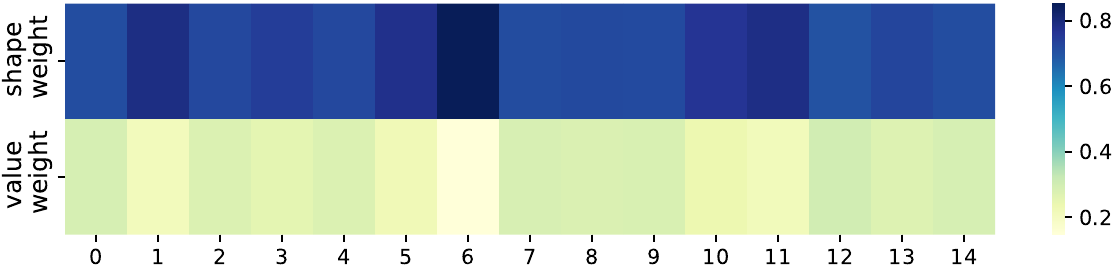}
    \caption{The heat map illustrates the distribution of the shape weight ($\lambda$) and value weight ($1-\lambda$) for each instance in the AtrialFibrillation dataset.}
    \label{fig:heatmap_shape}
\end{figure}

We justify the effectiveness of incorporating both shape and value through ablation study in Table ~\ref{ablation}. 
For Value only and Shape only, we remove the decision layer and only use the value and the shape branch, respectively. Table ~\ref{ablation} shows that VSFormer has the highest average accuracy and average rank, indicating the effectiveness of combining value and shape information for time series classification.

\subsection{Effectiveness Analysis}

We investigate the effectiveness of $\lambda$ in the decision layer. As described earlier, $\lambda$ in Formula \ref{eq:lambda_1} and \ref{eq:lambda_2} regulates the contribution of shape and value information in the final decision, allowing the model to emphasize the information that is more informative.

We test on the AtrialFibrillation dataset, which comprises two-channel ECG recordings, aiming to predict spontaneous termination of atrial fibrillation \cite{UEA}. The dataset has three classes: non-termination, self-terminating at least one minute after recording, and immediate termination within one second of recording end. \cite{shapenet} has highlighted the presence of discriminative patterns that can effectively distinguish between classes. This observation is corroborated by the learning of $\lambda$ in our model. As illustrated in the heat map in Figure \ref{fig:heatmap_shape}, the colors corresponding to the shape weights ($\lambda$) are notably darker than those for value weights ($1-\lambda$) across all 15 test set samples of AtrialFibrillation. This indicates that in every sample, the shape information carries more weight in determining the final classification results compared to the value information.

\begin{figure}[t]
\centering
\includegraphics[width=0.3\textwidth]{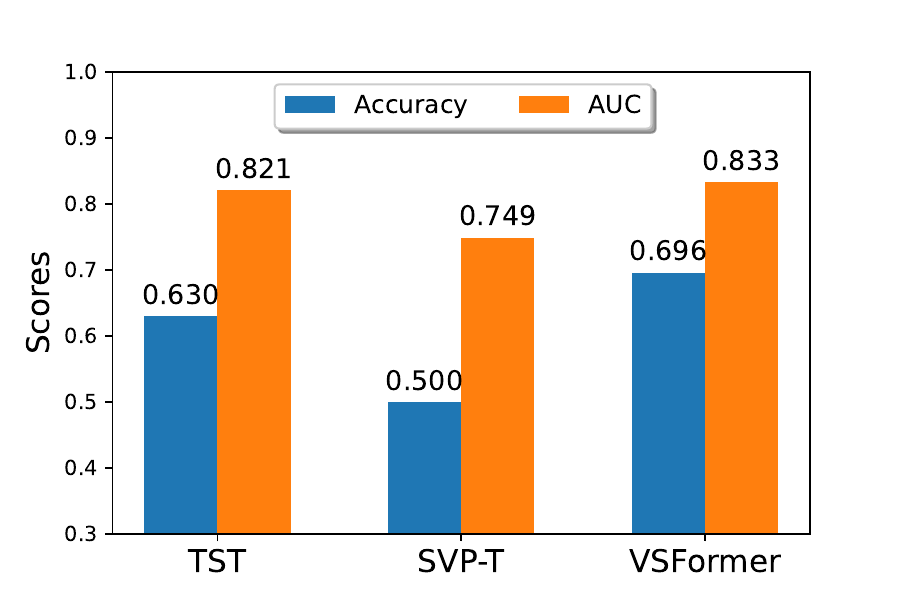} 
\caption{Performance comparison of TST, SVP-T, and VSFormer regarding the accuracy and AUC on the dataset.}
\label{fig:SF_results}
\vspace{-5mm}
\end{figure}

\begin{figure}[t]
    \centering
    \includegraphics[width=0.40\textwidth]{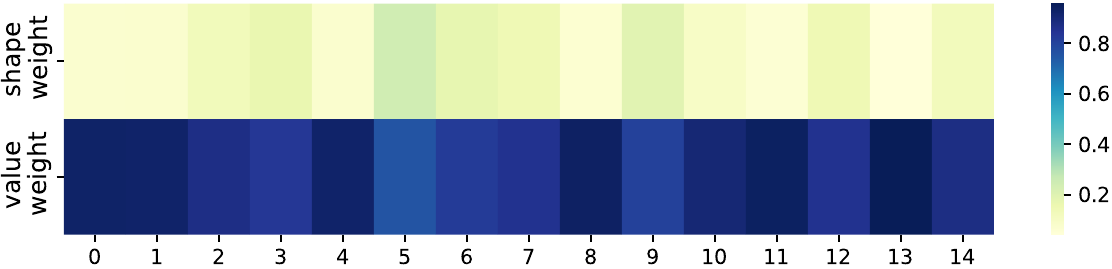}
    \caption{The heat map for 15 randomly selected test samples in the SWAN-SF dataset.}
    \label{fig:heatmap2}
\vspace{-3mm}
\end{figure}

\subsection{Case Study: Solar Flare Detection}\label{case_study}

To further illustrate the superiority of VSFormer in dealing with data that do not have clear discriminative patterns, we use the \textit{SWAN-SF} dataset, as discussed earlier. For data preprocessing, we employ linear interpolation to manage missing data and median smoothing to mitigate artificial spikes. Samples are extracted and categorized into four classes according to flare level using Sample Types sampling \cite{ajalvarez_thesis}. 
Each feature undergoes global Min-Max normalization. Since the dataset is extremely imbalanced, we implement undersampling to equalize sample numbers across classes. More details can be found in Appendix~\ref{ap:SF_dataset}.

Notably, most existing literature uses the timestamp-level method \cite{SF1,SF2,SF3,solarflare_tsf}, owing to the dataset's lack of discriminative patterns. We evaluate the performance of TST, SVP-T, and VSFormer on the dataset, and assess both accuracy and Area Under the Curve (AUC) metrics. 

As depicted in Figure \ref{fig:SF_results}, VSFormer outperforms the other two in both metrics, and TST outperforms SVP-T. The superior performance of TST over SVP-T can be attributed to its timestamp-level basis, while SVP-T relies on shape for time series representation—a challenge given the elusive nature of discriminative shapes in SWAN-SF data. VSFormer integrates both shape and value information, emphasizing value information as evidenced in Figure \ref{fig:heatmap2} (the color of the value weight is markedly darker than the shape weight), which results in better performance than SVP-T. Moreover, the benefits of TSI encoding and PESA in VSFormer over the learnable positional encoding and traditional self-attention employed by TST also explain VSFormer's superior performance.

\section{Conclusion}
In this work, we introduce VSFormer, a value and shape-aware Transformer tailored for MTSC. The model incorporates both discriminative patterns and numerical information, enhancing the performance in cases where discriminative patterns are lacking. In addition to using class-specific prior information to extend the encoding layer, we introduce it into self-attention learning to enhance classification-relevant features and reduce the impact of noise. Extensive experiments on all 30 UEA archives demonstrate the performance superiority of our model over SOTA models. The case study also shows that our model has superior performance in the case of the absence of discriminative patterns. 
In future work, we will focus on improving its efficiency on very large datasets and further streamlining the generated tokens.

\bibliographystyle{abbrv}
\bibliography{sdm25}

\clearpage
\appendix
\section{Appendix}
\subsection{Solar Flare Dataset}\label{ap:SF_dataset}
We use the \textit{Space Weather Analytics for Solar Flares (SWAN-SF)} dataset \cite{angryk_swan-sf_2020}, making use of 24 magnetic features with a 12-minute cadence derived from solar active region (AR) observations \cite{angryk_multivariate_2020} captured by the Helioseismic and Magnetic Imager (HMI) 
aboard NASA's Solar Dynamics Observatory. 
Flare events were captured by the National Oceanic and Atmospheric Administration
(NOAA) Geostationary Operational Environmental Satellites (GOES).  The samples generated are labeled as flaring (positive) targeting X and M class flares and non-flaring (negative) for smaller C and B flares as well as when no flares were present (N). Each sample is 24 hours long and includes a sequence of 120 12-minute observations.  

To capture a broad set of photospherical observations providing a diverse set of sample time distributions, we use Sample Types sampling \cite{ajalvarez_thesis} to extract samples within $\pm60^\circ$ from the solar central meridian in 4 different ways. One sample type, Flaring Active Region Flaring Observations (FAF), includes observations leading to a target (X or M class) flare event. These types of samples make up the positive class, whereas the other three make up the negative class.  The second sample type, Flaring Active Region Non-flaring Observations (FAN), is also taken from ARs containing target flare events, but these observations do not lead to a flare event providing us with a different (non-flaring) perspective of the same AR. The third sample type, Non-flaring Active Region Flaring Observations (NAF), leads to a (smaller) non-target (C or B class) flare event considered non-flaring. The fourth and last type of sample, Non-flaring Active Region Non-flaring Observations (NAN), is taken from non-flaring active regions at different points in time of the given ARs. The dataset is extremely imbalanced; therefore, we perform undersampling, producing 114 samples for each class.

\begin{table}[!h]
\caption{Hyperparameters for each dataset}
\begin{adjustbox}{width=0.48\textwidth,center}
\begin{tabular}{@{}lcccc@{}}
\toprule
                          & \multicolumn{1}{l}{bs} & \multicolumn{1}{l}{d\_model} & \multicolumn{1}{l}{d\_feedforward} & \multicolumn{1}{l}{lr} \\ \midrule
ArticularyWordRecognition & 8                      & 32                           & 128                                & 1e-3                   \\
AtrialFibrillation        & 8                      & 8                           & 64                                 & 1e-3                   \\
BasicMotions              & 8                      & 8                            & 64                                 & 1e-3                   \\
CharacterTrajectories     & 16                     & 16                            & 32                                 & 1e-4                   \\
Cricket                   & 8                      & 8                           & 32                                 & 1e-3                   \\
DuckDuckGeese             & 4                      & 8                           & 16                                 & 1e-3                   \\
EigenWorms                & 8                      & 8                            & 16                                 & 1e-4                   \\
Epilepsy                  & 8                     & 16                           & 32                                 & 1e-3                   \\
ERing                     & 8                      & 64                            & 128                                 & 1e-3                   \\
EthanolConcentration      & 8                     & 16                           & 64                                 & 1e-3                   \\
FaceDetection             & 8                      & 64                           & 64                                 & 1e-3                   \\
FingerMovements           & 16                      & 8                           & 128                                 & 1e-3                   \\
HandMovementDirection     & 16                      & 8                           & 32                                 & 1e-4                   \\
Handwriting               & 4                      & 16                           & 128                                 & 1e-3                   \\
Heartbeat                 & 8                      & 64                           & 64                                 & 1e-3                   \\
InsectWingbeat            & 16                     & 16                           & 32                                 & 1e-3                   \\
JapaneseVowels            & 8                      & 64                           & 16                                 & 1e-4                   \\
Libras                    & 8                     & 64                           & 128                                & 1e-4                   \\
LSST                      & 8                     & 64                           & 32                                 & 1e-4                   \\
MotorImagery              & 16                     & 32                           & 32                                 & 1e-4                   \\
NATOPS                    & 16                      & 64                           & 16                               & 1e-3                  \\
PenDigits                 & 32                     & 16                           & 128                                 & 1e-4                   \\
PEMS-SF                   & 8                      & 64                          & 16                                & 1e-3                   \\
Phoneme                   & 32                     & 64                           & 16                                 & 1e-4                   \\
RacketSports              & 16                     & 64                           & 64                                 & 1e-3                   \\
SelfRegulationSCP1        & 8                      & 8                            & 64                                 & 1e-3                   \\
SelfRegulationSCP2        & 16                     & 8                            & 64                                 & 1e-3                   \\
SpokenArabicDigits        & 32                     & 16                           & 32                                 & 1e-3                   \\
StandWalkJump             & 8                      & 16                           & 128                                & 1e-4                   \\
UWaveGestureLibrary       & 32                     & 32                           & 32                                 & 1e-4                   \\
SolarFlare                & 4                      & 16                           & 128                                & 1e-4                   \\ \bottomrule
\end{tabular}
\end{adjustbox}
\label{tab:hyper}
\end{table}

\subsection{Hyperparameters}\label{ap:hyper}
Here, we provide hyperparameters used in our model. Table \ref{tab:hyper} shows the batch size (bs), the dimension of the input for the encoder in the shape branch (d\_model), the dimension of the feedforward layer in the shape branch (d\_feedforward), and the learning rate (lr) for each dataset. Since the input token size of the value branch is uniform, we fix the d\_model of the value branch to 8 and the d\_feedforward to 16. For both shape and value branch, we fix the number of heads 
to 8 and the number of encoder layers to 1. We adopt batch normalization, instead of layer normalization, since it could provide better performance on time series applications \cite{tst}. 

\subsection{Sensitivity Analysis}
Here, we provide the sensitivity analysis for $k$ in shape token generation and $M$ in value token generation. \\

\begin{figure}[t]
    \centering
    \begin{subfigure}[b]{0.45\textwidth}
        \includegraphics[width=\textwidth]{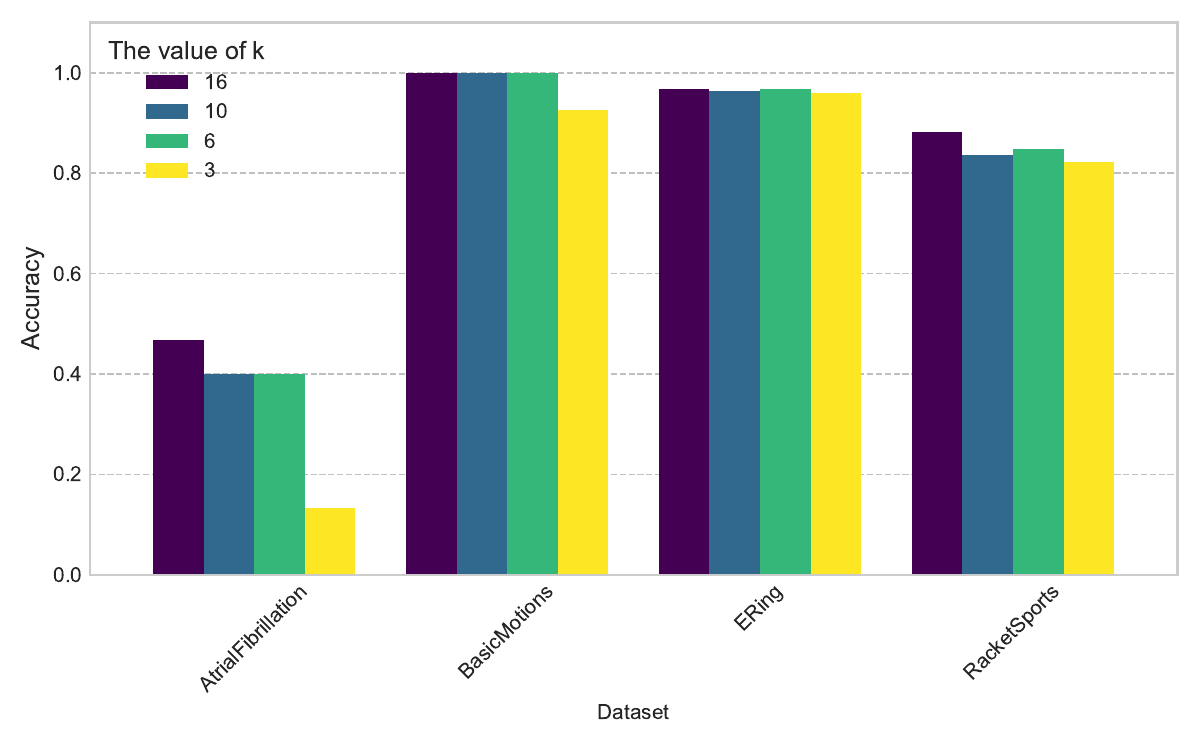}
        \caption{Accuracy for various datasets with different $k$ (16, 10, 6, 3).}
        \label{fig:hyper_1}
    \end{subfigure}
    \hfill
    \begin{subfigure}[b]{0.45\textwidth}
        \includegraphics[width=\textwidth]{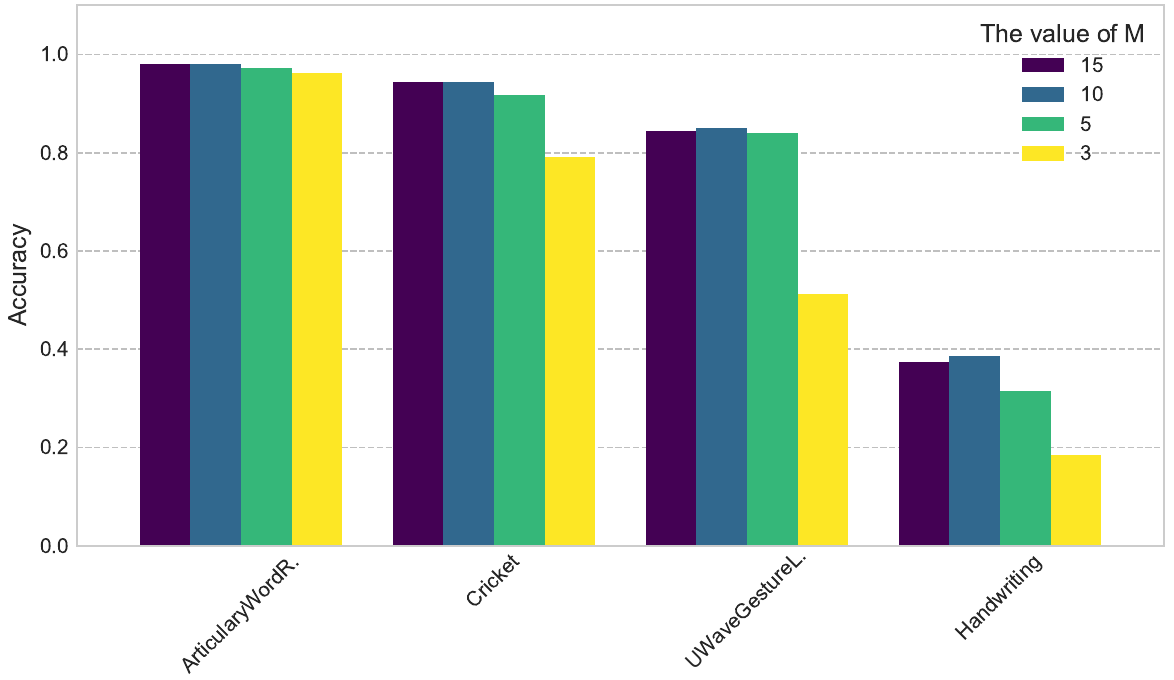}
        \caption{Accuracy for various datasets with different $M$ (15, 10, 5, 3).}
        \label{fig:hyper_2}
    \end{subfigure}
    \caption{Comparison of accuracy across different datasets and selection for $k$ and $M$.}
    \label{fig:hyper_accuracy}
\end{figure}

\noindent\textbf{Selection for $k$.}
$k$ represents the number of top motif pairs for each variable and class. To determine the optimal value of $k$, we test four datasets using only the shape branch to eliminate the impact of the value branch. As illustrated in Fig.~\ref{fig:hyper_1}, setting $k = 6$ generally achieves good results, while for certain datasets such as AtrialFibrillation and RacketSports, $k = 16$ performs better. 
\\

\noindent\textbf{Selection for $M$.}
$M$ denotes the maximum number of equal-length intervals into which each time series is segmented. We test four datasets using only the value branch to eliminate the influence of the shape branch. As shown in Fig.~\ref{fig:hyper_2}, $M = 10$ achieves the best results across all datasets. Further increasing $M$ does not improve accuracy. 


\subsection{Interpreting Results}

\begin{figure*}[t]
\centering
    \includegraphics[width=1\textwidth]{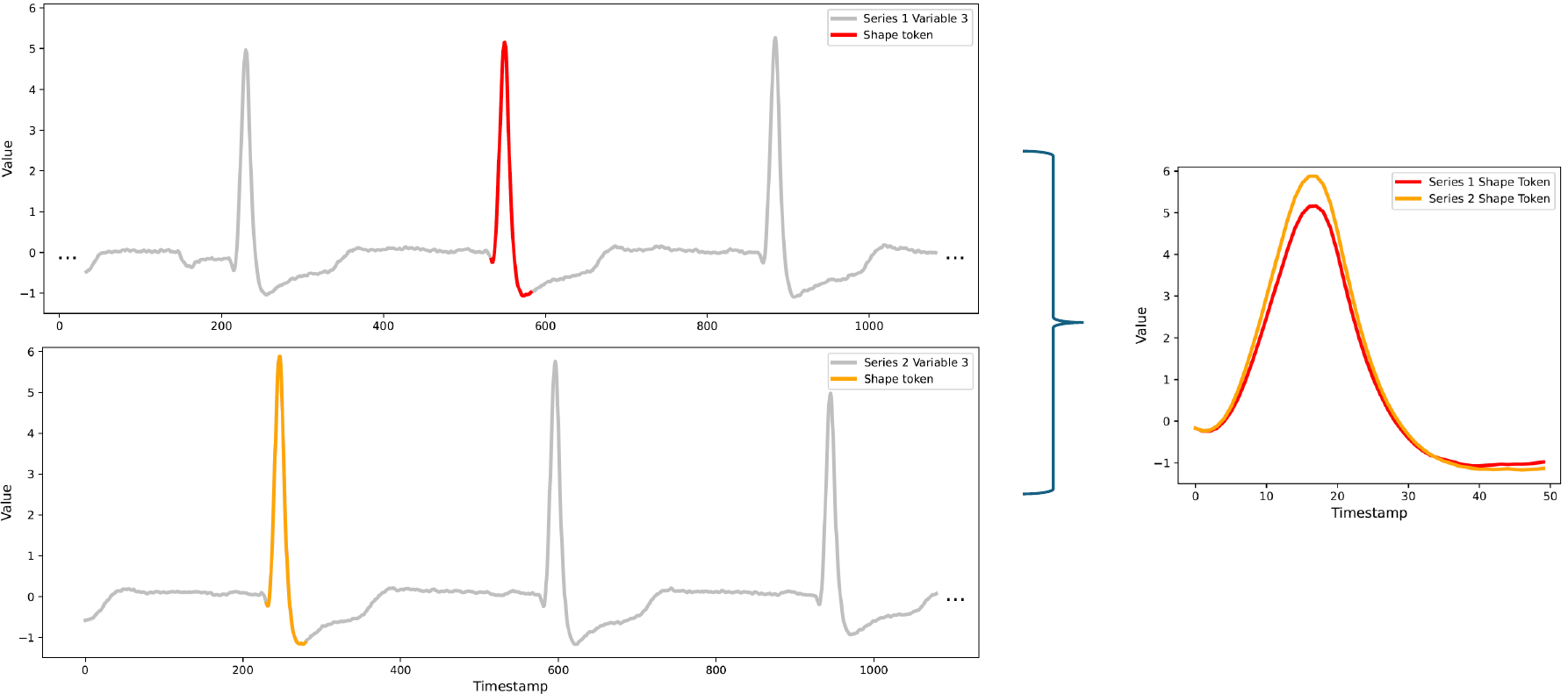}
\caption{Visualization for the StandWalkJump dataset, showing time series for the third variable of each instance and the comparison for two shape tokens. The top left figure represents a "standing" class instance. The bottom left figure shows a "jumping" class instance. Shape tokens with the highest weights are highlighted in red and orange, respectively. The right figure shows the comparison between two representative shape tokens from two different classes.}
\label{fig:shape_token_vis}
\end{figure*}

\begin{figure*}[t]
\centering
\begin{subfigure}[]{0.46\textwidth}
    \includegraphics[width=\textwidth]{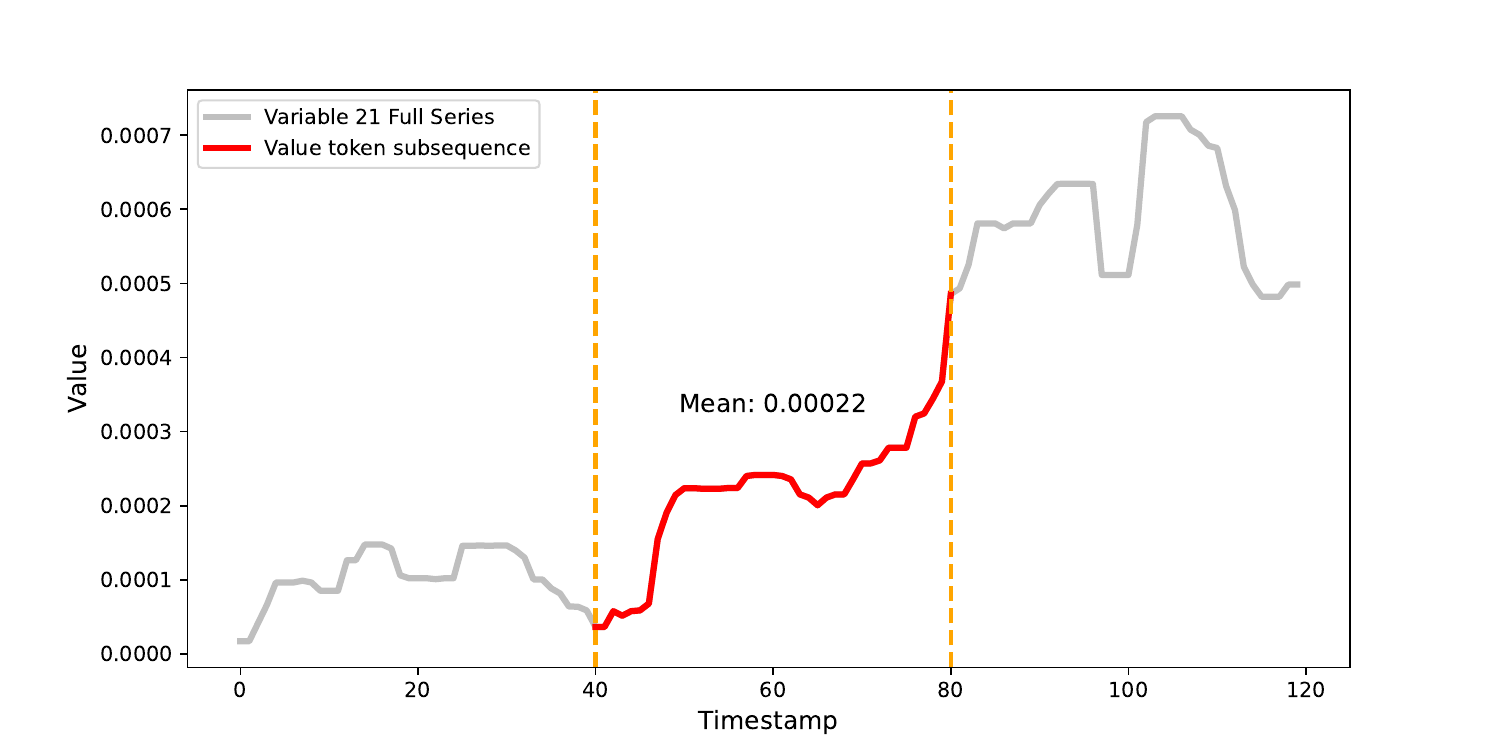}
\end{subfigure}
\hspace{-0.5pt}
\begin{subfigure}[]{0.455\textwidth}
    \includegraphics[width=\textwidth]{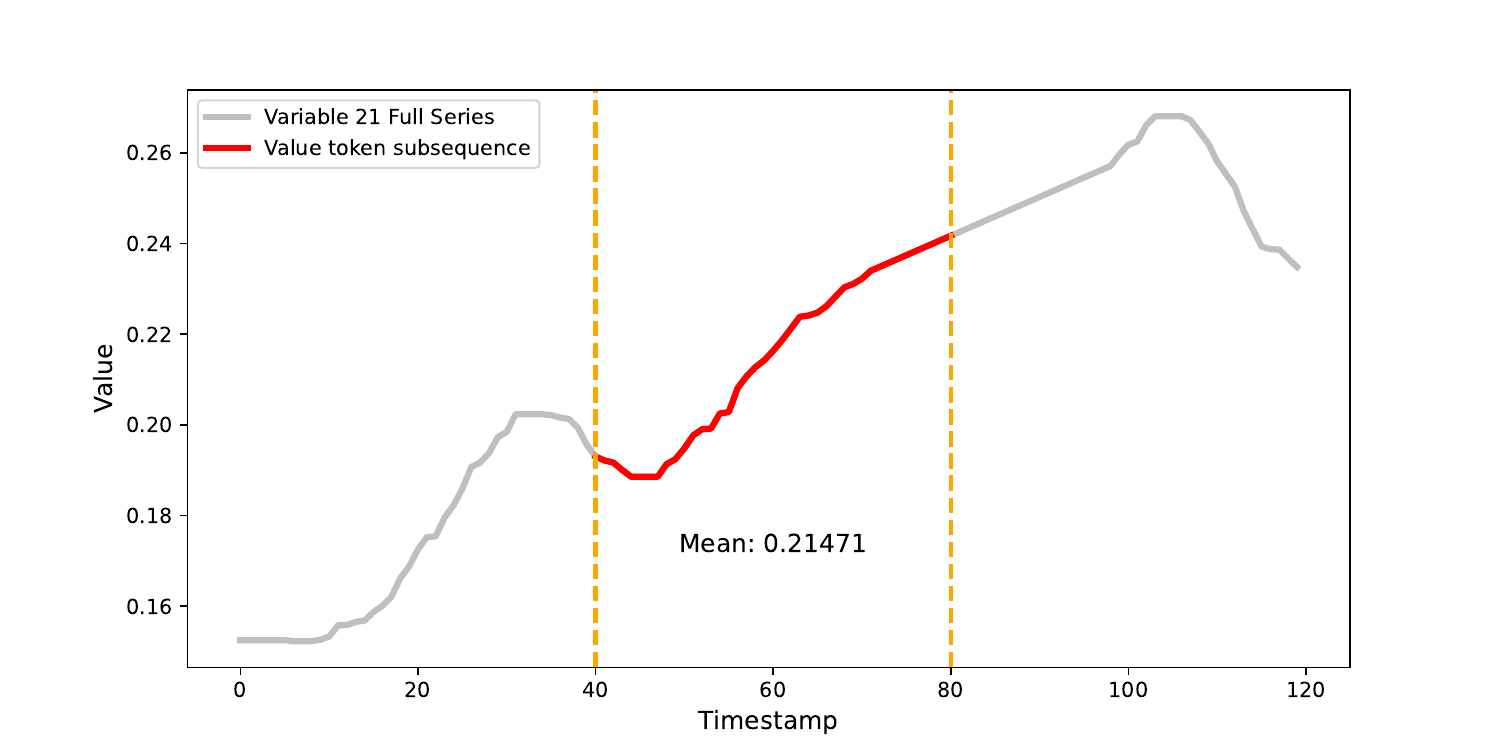}
\end{subfigure}
\caption{Visualization of two instances from the SWAN-SF dataset showing time series for the 21st variable of each instance. The left figure represents a "FAF" (positive/flaring) class instance. The right figure shows a "NAF" (negative/non-flaring) class instance. The subsequence within the interval represented by the value token with the highest weight is highlighted in red. Both value tokens are represented by the mean value within the interval.}
\label{fig:value_token_vis}
\end{figure*}

Given that our model utilizes subsequences (shape tokens) and statistical values of intervals (value tokens) as input, it shows excellent interpretability. We use the StandWalkJump and SWAN-SF datasets to illustrate the interpretability of the shape and value information, respectively. The former dataset presents discriminative patterns, whereas the latter lacks such patterns but has discriminative value information as described in Section~\ref{intro} and~\ref{case_study}. We follow \cite{SVPT} to calculate attention and highlight those input tokens with the highest weight in the results.

Fig.~\ref{fig:shape_token_vis} presents the representative shape tokens for different classes in the StandWalkJump dataset. The dataset contains three classes, standing, walking, and jumping, representing short-duration ECG signals under different physical activities~\cite{UEA}. The upper left figure shows the time series of the third variable for an instance of the "standing" class, while the bottom left figure shows the same variable for an instance of the "jumping" class. We highlight the shape tokens with the highest weights for both instances in red and orange, respectively, and compare them in the right figure. While the differences are subtle, we observe that the shape token for the “jumping” class is sharper (exhibiting a higher peak) compared to the shape token for the “standing” class.

Fig.~\ref{fig:value_token_vis} illustrates the representative value tokens for the different classes in the SWAN-SF dataset. As discussed in Section~\ref{case_study}, the dataset contains four classes: FAF, FAN, NAF, and NAN. Each instance is made up of 24 different magnetic time series variables. The left figure corresponds to an instance of the "FAF" class, whilst the right figure corresponds to an instance of the "NAF" class. The value token with the highest weight is highlighted in red, indicating the value information within such an interval that can significantly impact the model's decision-making. The value tokens represent the mean value within the highlighted interval for both instances. There is an obvious difference in the values within the same interval between the two instances with the value of the "NAF" instance (0.21471) being three orders of magnitude higher than that of the "FAF" instance (0.00022).

\subsection{Handling 'N/A' results}
When handling the accuracy results of 'N/A' in Table ~\ref{results}, we follow \cite{SVPT} by assigning them to 0. This can significantly influence the average accuracy due to its sensitivity to assigned values. To be fair, we decide not to include it in Table ~\ref{results}. However, this will not affect the average rank since it is determined by relative positions rather than the actual accuracy. Therefore, we retain the average rank as a metric in our results. We also tried throwing away the results of 'N/A' by excluding them from the average rank. The results showed that our method still has the highest average rank.

In Table ~\ref{ablation}, we include both average accuracy and average rank, as there is no 'N/A' in the results.

\subsection{Fair Comparison}
For a fair comparison, we consider only the non-pretrained baseline models. Consequently, results from TS-TCC \cite{TSTCC} and TARNet \cite{TARNet} are excluded as they rely on pre-training, which falls outside our current scope.

\end{document}